\documentclass[11pt,letterpaper]{article}
\usepackage[letterpaper]{geometry}
\usepackage{acl2012}
\usepackage{booktabs}
\usepackage{times}
\usepackage{latexsym}
\usepackage{amsmath}
\usepackage{amsfonts}
\usepackage{graphicx, subfigure}
\usepackage{color}
\usepackage{array}
\usepackage{tabulary}
\newcolumntype{K}[1]{>{\centering\arraybackslash}p{#1}}
\makeatletter
\newcommand{\@BIBLABEL}{\@emptybiblabel}
\newcommand{\@emptybiblabel}[1]{}
\makeatother
\usepackage[hidelinks]{hyperref}

\title{Conversation Modeling on Reddit using a Graph-Structured LSTM}

\author{Vicky Zayats \\
  Electrical Engineering Department \\
  University of Washington \\
  {\tt vzayats@uw.edu} \\\And
  Mari Ostendorf \\
  Electrical Engineering Department \\
  University of Washington \\
  {\tt ostendor@uw.edu} \\}

\date{}

\begin{document}

\maketitle

\begin{abstract}
This paper presents a novel approach for modeling threaded discussions on social media using a graph-structured bidirectional LSTM which represents both hierarchical and temporal conversation structure. In experiments with a task of predicting popularity of comments in Reddit discussions, the proposed model outperforms a node-independent architecture for different sets of input features. Analyses show a benefit to the model over the full course of the discussion, improving detection in both early and late stages. Further, the use of language cues with the bidirectional tree state updates helps with identifying controversial comments.
\end{abstract}

\section{Introduction}

\begin{figure}[t]
\centering
\includegraphics[scale=0.09]{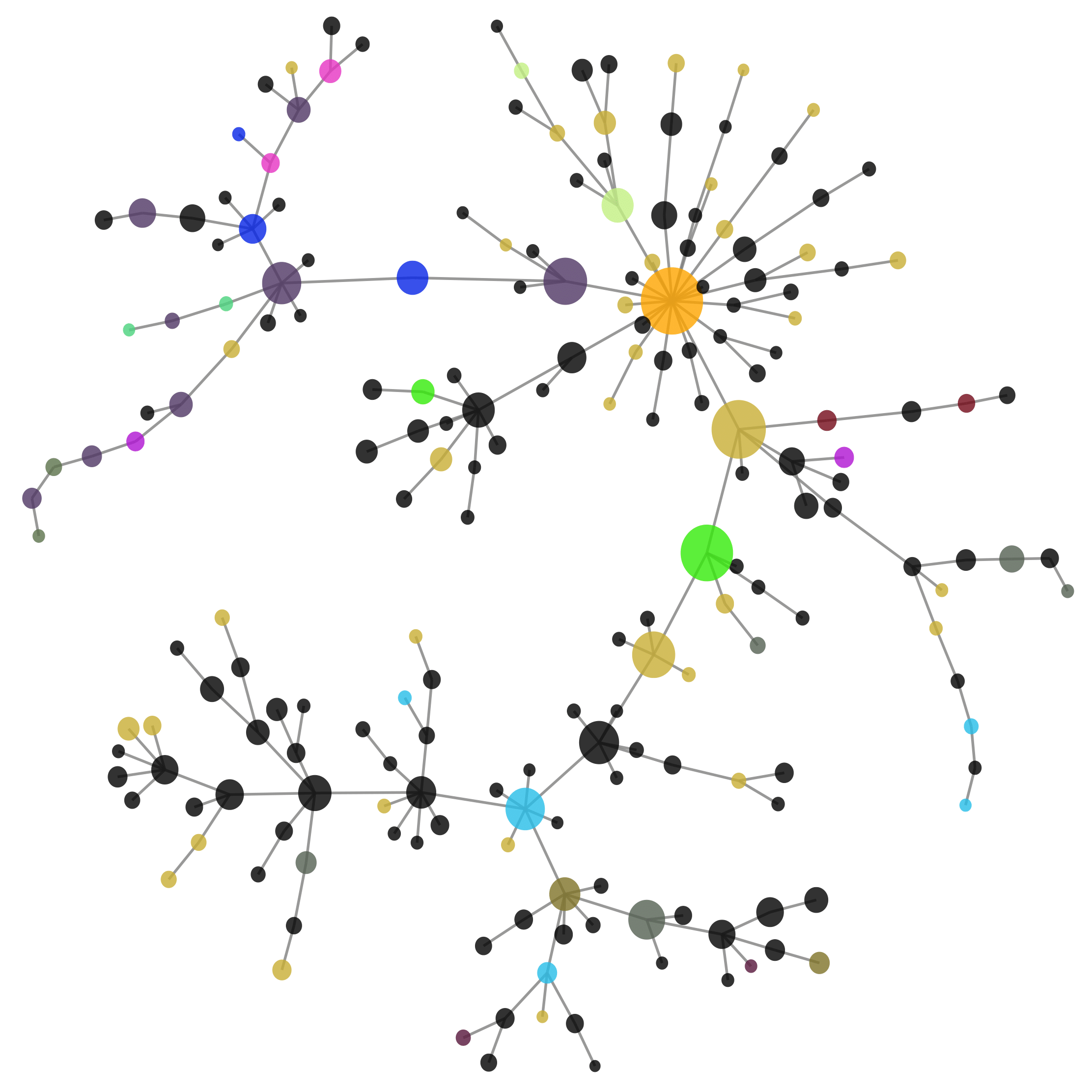}
\caption{Visualization of a sample thread on Reddit.}
\label{fig: visual}
\end{figure}

Social media provides a convenient and widely used platform for discussions among users. When the comment-response links are preserved, those conversations can be represented in a tree structure where comments represent nodes, the root is the original post, and each new reply to a previous comment is added as a child of that comment. Some examples of popular services with tree-like structures include Facebook, Reddit, Quora, and StackExchange. Figure \ref{fig: visual} shows an example conversation on Reddit, where bigger nodes indicate higher upvoting of a comment.\footnote{The tool https://whichlight.github.io/reddit-network-vis was used to obtain this visualization.} In services like Twitter, tweets and their retweets can also be viewed as forming a tree structure. When time stamps are available with a contribution, the nodes of the tree can be ordered and annotated with that information. The tree structure is useful for seeing how a discussion unfolds into different subtopics and showing differences in the level of activity in different branches of the discussion.

\begin{figure*}[t]
\centering
\subfigure[Forward hierarchical and timing structure]{\label{fig:a}
\includegraphics[scale=0.5]{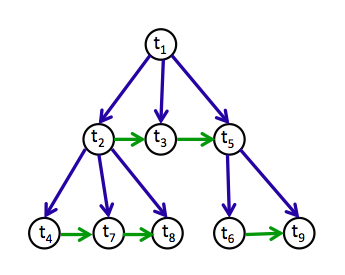}} \qquad
\subfigure[Backward hierarchical and timing structure]{\label{fig:b}
\includegraphics[scale=0.5]{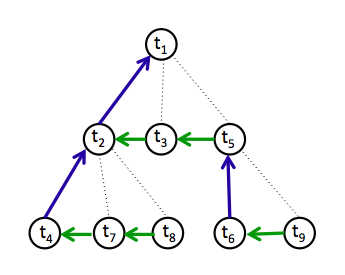}} \qquad
\caption{An example of model propagation in a graph-structured LSTM. Here, the node name are shown in a chronological order, e.g. comment $t1$ was made earlier than $t2$. \ref{fig:a} Propagation of graph-structured LSTM in the forward direction. Blue arrows represent hierarchical propagation, green arrows represent timing propagation.\ref{fig:b} Backward hierarchical (blue) and timing (green) propagation of graph-LSTM. \label{lstm_example}}
\end{figure*}

Predicting popularity of comments in social media is a task of growing interest. Popularity has been defined in terms of the volume of the response, but when the social media platform has a mechanism for readers to like or dislike comments (or, upvote/downvote), then the difference in positive/negative votes provides a more informative score for popularity prediction. This definition of popularity, which has also been called community endorsement \cite{fang2016learning}, is the task of interest in our work on tree-structured modeling of discussions.

Previous studies found that the time when the comment/post was published has a big impact on its popularity \cite{lakkaraju2013s}. In addition, the number of immediate responses can be predictive of the popularity, but some comments with a high number of replies can be either controversial or have a highly negative score. Language should be extremely important for distinguishing these cases. Indeed, community style matching is shown to be correlated to comment popularity in Reddit \cite{TranOst2016}. However, learning useful language cues can be difficult due to the low frequency of these events and the dominance of time, topic and other factors. Thus, in several prior studies, authors constrained the problem to reduce the effect of those factors \cite{lakkaraju2013s,tan2014effect,jaech2015talking}. In this study, we have no such constraints, but attempt to use the tree structure to capture the flow of information in order to better model the context in which a comment is submitted, including both the history it responds to as well as the subsequent response to that comment. 

To capture discussion dynamics, we introduce a novel approach to modeling the discussion using a bidirectional graph-structured LSTM, where each comment in the tree corresponds to a single LSTM unit. In one direction, we capture the prior history of contributions leading up to a node, and in the other, we characterize the response to that comment. Motivated by prior findings that both response structure and timing are important in predicting popularity \cite{fang2016learning}, the LSTM units include both hierachical and temporal components to the update, which distinguishes this work from prior tree-structured LSTM models. We assess the utility of the model in experiments on popularity prediction with Reddit discussions, comparing to a neural network baseline that treats comments independently but leverages information about the graph context and timing of the comment. We analyze the results to show that the graph LSTM provides a useful summary representation of the language context of the comment.

As in \cite{fang2016learning}, but unlike other work \cite{He2016}, our model makes use of the full discussion thread in predicting popularity. While knowledge of the full discussion is only useful for post-hoc analysis of past discussions, it is reasonable to consider initial responses to a comment, particularly given that many responses occur within minutes of someone posting a comment.  Comments are often popular because of witty analogies made, which requires knowledge of the world beyond what is captured in current models. Responses to these comments, as well as to controversial comments, can improve popularity prediction. Responses of others clearly influence the likelihood of someone to like or dislike a comment, but also whether they even read a comment. By introducing a forward-backward tree-structured model, we provide a mechanism for leveraging early responses in predicting popularity, as well as  a framework for better understanding the relative importance of these responses.

The main contributions of this paper include: a novel approach for representing tree-structured language processes (e.g., social media discussions) with LSTMs; evaluation of the model on the popularity prediction task using Reddit discussions; and analysis of the performance gains, particularly with respect to the role of language context.

\section{Method}
\label{method}

The proposed model is a bidirectional graph LSTM that characterizes a full threaded discussion, assuming a tree-structured response network and accounting for the relative order of the comments. Each comment in a conversation corresponds to a node in the tree, where its parent is the comment that it is responding to and its children are the responding comments that it spurs ordered in time. 
Each node in the tree is represented with a single recurrent neural network (RNN) unit that outputs a vector (embedding) that characterizes the interim state of the discussion, analogous to the vector output of an RNN unit which characterizes the word history in a sentence. In the forward direction, the state vector can be thought of as a summary of the discussion pursued in a particular branch of the tree, while in the backward direction the state vector summarizes the full response subtree that followed a particular comment. The state vectors for the forward and backward directions are concatenated for the purpose of predicting comment karma. The RNN updates -- both forward and backward -- incorporate both temporal and hierarchical (tree-structured) dependencies, since commenters typically consider what has already been said in response to a parent comment. Hence, we refer to it as a graph-structured RNN rather than a tree-structured RNN. 
Figures \ref{fig:a} and \ref{fig:b} show an example of the state connections associated with hierarchical and timing structures for the forward and backward RNNs, respectively. 


The supervision signal in training will impact the character of the state vector, and the forward and backward state sub-vectors are likely to capture different phenomena. Here, the objective is to predict quantized comment karma. We anticipate that the forward state will capture relevance and informativeness of the comment, and the backward process will capture sentiment and richness of the ensuing discussion.

The specific form of the RNN used in this work is an LSTM. The detailed implementation of the model is described in the sections to follow.

\subsection{Graph-structured LSTM}

Each node in the tree is associated with an LSTM unit.
The input $x_t$ is an embedding that can incorporate both comment text and local submission context features associated with thread structure and timing, described further in section~\ref{sec:input}. The node state vector $h_t$ is generated using a modification of the standard LSTM equations to include both hierarchical and timing structures for each comment. Specifically, we use two forget gates - one for the previous (or subsequent) hierarchical layer, and one for the previous (or subsequent) timing layer. 
 
In order to describe the update equations, we introduce notation for the hierarchical and timing structure. In Figure~\ref{lstm_example}, the nodes in the tree are numbered in the order that the comments are contributed in time. To characterize graph structure, let $\pi(t)$ denote the parent of $t$ and $\kappa(t)$ its first child. Time structure is represented only among a set of siblings: $p(t)$ is the sibling predecessor in time, and $s(t)$ is the sibling successor. The pointers $\kappa(t)$, $p(t)$ and $s(t)$ are set to $\emptyset$ when $t$ has no child, predecessor, or successor, respectively. For example, in Figure~\ref{fig:a}, the node $t_2$ will have $\pi(t_2)=t_1$, $\kappa(t_2)=t_4$, $p(t_2)=\emptyset$ and $s(t_2)=t_3$, and the node $t_3$ will have $\pi(t_3)=t_1$, $\kappa(t_3)=\emptyset$, $p(t_3)=t_2$ and $s(t_3)=t_5$. 

Below we provide the update equations for the forward process, using the subscripts $i,\ f,\ g,\ c,$ and $o$ for the input gate, temporal forget gate, hierarchichal forget gate, cell, and output, respectively. The vectors $i_t$, $f_t$, and $g_t$ are the weights for new information, remembering old information from siblings, and remembering old information from the parent, respectively.  $\sigma$ is a sigmoid function, and $\circ$ indicates the Hadamard product. 
If $p(t)=\emptyset$, then $h_{p(t)}$ and $c_{p(t)}$ are set to the initial state value.
\begin{equation}
\label{eq: 1}
\begin{aligned}
i_{t} &= \sigma(W_{i}x_{t} + U_{i}h_{p(t)}  + V_{i}h_{\pi(t)} + b_{i}) \\
f_t &= \sigma(W_{f}x_{t} + U_{f}h_{p(t)} + V_{f}h_{\pi(t)} + b_{f}) \\
g_t &= \sigma(W_{g}x_{t} + U_{g}h_{p(t)} + V_{g}h_{\pi(t)} + b_{g}) \\
\tilde{c}_{t} &= W_{c}x_{t} + U_{c}h_{p(t)} + V_{c}h_{\pi(t)} + b_{c} \\
c_t &= f_t \circ c_{p(t)} + g_t \circ c_{\pi(t)} + i_t \circ \tilde{c}_t \\
o_t &= \sigma(W_{o}x_{t} + U_{o}h_{p(t)} +  V_{o}h_{\pi(t)} + b_{o}) \\
h_t &= o_t \circ tanh( c_t)
\nonumber
\end{aligned}
\end{equation}

When the whole tree structure is known, we can take advantage of the full response subtree to better represent the node state.  To that end, we define a backward LSTM that has a similar set of update equations except that only the first child  will pass the hidden state to its parent. Specifically, the update equations are the same except that $\pi(t)$ is replaced with $\kappa(t)$, $p(t)$ is replaced with $s(t)$, and a different set of weight matrices and bias vectors are learned.

Let $+$ and $-$ indicate forward and backward embeddings respectively. 
On top of the LSTM unit, the forward and backward state vectors are concatenated and passed to a softmax layer to predict 8 quantized karma levels:
\begin{equation}
\label{eq: 2}
\begin{aligned}
P(y_t=j|x,h) = \frac{\exp(W_s^j[h^+_t; h^-_t])}{\sum_{k=1}^{8}\exp(W_s^k[h^+_t; h^-_t])}
\nonumber
\end{aligned}
\end{equation}
where $x$ and $h$ correspond to the set of input features and state vectors (respectively) for all nodes in the discussion.


\subsection{Input Features}
\label{sec:input}

The full model includes two types of features in the input vector, including non-textual features associated with the submission context and the textual features of the comment at that node.

The {\bf submission context} features are extracted from the graph and metadata associated with the comment, motivated by prior work showing that context factors such as the forum, timing and author of a post are very useful in predicting popularity. The submission context features include:
\begin{itemize}
\item \textit{Timing}: time since root, time since parent (in hours), number of later comments, and number of previous comments
\item \textit{Author}: a binary indicator as to whether the author is the original poster, and number of comments made by the author in the conversation
\item \textit{Graph-location}: depth of the comment (distance from the root), and number of siblings
%
\item \textit{Graph-response}:  number of children (direct replies to the comment), height of the subtree rooted from the node, size of that subtree, number of children normalized for each thread (2 normalization techniques), subtree size normalized for each thread (2 normalization techniques). 
\end{itemize}
Two methods are used to normalize the subtree size and number of children to compensate for variation associated with the size of the discussion, specifically: i)~subtract the mean feature value in the thread, and ii)~divide by the square root of the rank of the feature value in the thread.

These features are a superset of those used in \cite{fang2016learning}.
The subvector including all these features is denoted $x_t^s$.


The {\bf comment text} features, denoted $x_t^c$, are generated using a simple average bag-of-words representation learned during the training: 
\begin{equation}
\label{eq:comment}
\begin{aligned}
x_t^c = \frac{1}{N}\sum_{i=1}^{N}W_e^i
\nonumber
\end{aligned}
\end{equation}
where
$W_e^i$ is an embedding of the $i$-th word in the comment, and $N$ is the number of words in the comment. 
Comments longer than 100 words were truncated to reduce noise associated with long comments, assuming that the early portion carries the most information. 
The percentage of the comments that exceed 100 words is around $11\%-14\%$ for the subreddits used in the study. 
 In all experiments, the word embedding dimension is $d=100$, and the vocabulary includes only words that occurred at least 10 times in the dataset.

The input vector $x_t$ is set to either $x_t^s$ or $[x_t^s; x_t^c]$, depending on whether the experiment uses text.

\subsection{Pruning}
\label{sec:prune}


Often the number of comments in a single subtree can be large, which leads to high training costs. A large percentage of the comments are low karma and minimally relevant for predicting karma of neighbors, and many can be easily identified with simple graph and timing features (e.g. having no replies or contributed late in the discussion). Therefore, we introduce a preprocessing step that identifies comments that are highly likely to be low karma to decrease the computation cost. We then assign these nodes to be level 0 and prune them out of the tree, but retain a count of nodes pruned for use in a count-weighted bias term in the update to capture information about response volume.

For detecting low karma comments, we train a simple SVM classifier to identify comments at the 0 karma level
based on the submission context features. If a pruned comment leads to a disconnected graph (e.g., an internal node is pruned but not its children), then the comment is retained in the tree. 
In testing, all pruned comments are given a predicted level of 0 and accounted for in the evaluation. 

The state updates have an additional bias term for any nodes that have subsequent sibling or children comments pruned. For example, consider Figure~\ref{lstm_example}, if nodes $\{t_5, t_6, t_7, t_9\}$ are pruned, then $t_8$ will have a modified forward update, and
$t_3, t_4$ will have a modified backwards update.
At node $t$, define $M_t^{\kappa}$ to be the number of levels pruned below it, $M_t^p$ as the number of immediately preceeding comments pruned in its subgroup (responding to the same parent), and $M_t^s$ as the number of subsequent comments pruned in its subgroup plus the non-initial comments in the associated subtrees. In the example above, 
$M_3^{\kappa}=1,\ M_3^s=2,\ M_4^s=1,\ M_8^p=1$, and all other $M_t^*=0$. The pointers are updated reflect the structure of the pruned tree, so $p(8)=4,\ s(4)=8,\ s(3)=\emptyset$.
The bias vectors $r_{\kappa}$, $r_p$ and $r_s$ are associated with the different sets of nodes pruned. 

Let $+$ and $-$ indicate forward and backward embeddings respectively. The forward update has an adjusted predecessor contribution 
$(h^+_{p(t)}+ M_t^p r_p).$ The backward update adds $M_t^s r_s+M_t^{\kappa} r_{\kappa}$ to either $h^-_{s(t)}$ or $h^-_{\kappa(t)}$, depending on whether it is a time or hierarchical update, respectively.


\subsection{Training}

The objective function is minimum cross-entropy over the quantized levels. All model parameters are jointly trained using the adadelta optimization algorithm \cite{adadelta}. Word embeddings are initialized using word2vec skip-gram embeddings \cite{word2vec} trained on all
comments from the corresponding subreddit. The code is implemented in Theano \cite{theano}
and is available at (github URL omitted for blind reviewing).
%
%
We tune the model over different dimensions of the LSTM unit, and use the performance on the development set as a stopping criteria for the training. 

\section{Experiments}
\label{sec:experiments}

\subsection{Data}

Reddit\footnote{reddit.com} is a popular discussion forum platform consisting of a large number of subreddits focusing on different topics and interests. In our study, we experimented with 3 subreddits: askwomen, askmen, and politics. All the data consists of discussions made in the period between January
1, 2014 and January 31, 2015.  Table ~\ref{tab:data} shows the total amount of data used for each of the subreddits. For each subreddit, the threads were randomly distributed between training, development (dev) and test sets with the proportions of 6:2:2. The performance of the pruning classifier on the dev set is presented in Table ~\ref{tab:prune}.

\begin{table}
\begin{center}
\begin{tabular}{|l|c|c|c|c|}
\hline \bf subreddit  & \bf comments & \bf threads & \bf vocab size\\ \hline
askwomen & 0.8M & 3.5K & 32K \\ \hline
askmen & 1.1M & 4.5K & 35K \\ \hline
politics & 2.2M & 4.9K  & 55K \\ \hline
\end{tabular}
\caption{\label{tab:data} Data statistics.}
\end{center}
\end{table} 

\begin{table}
\begin{center}
\begin{tabular}{|l|c|c|c|}
\hline \bf subreddit  &  Prec & Rec & \% pruned \\ \hline
askwomen & 67.9 & 72.4 & 36.9\\ \hline
askmen & 60.1 & 75.3 & 36.1 \\ \hline
politics  & 49.6 & 60.3 & 47.5 \\ \hline
\end{tabular}
\caption{Precision and recall of the pruning classifier and percentage of comments pruned.}
\label{tab:prune}
\end{center}
\end{table}



\subsection{Task and evaluation metrics}
\label{sec: levels}
Reddit karma has a Zipfian distribution, highly skewed toward the low-karma comments. Since the rare high karma comments are of greatest interest in popularity prediction, \cite{fang2016learning} proposes a task of predicting quantized karma (using a nonlinear head-tail break rule for binning) with evaluation using a macro average of the F1 scores for predicting whether a comment exceeds each different level. Experiments reported here use this framework. 

Specifically, all the comments with karma lower than 1 are assigned to level 0, and each subsequent level corresponds to karma less than or equal to the median karma in the rest of the comments based on the training data statistics. Each subreddit has 8 quantized karma levels based on its karma distribution. There are 7 binary subtasks (does the comment have karma at level $j$ or higher for $j=1,\ldots , 7$), and the scoring metric is the macro average of $F1(j)$. For tuning hyperparameters and as a stopping criterion, we use a linearly weighted average of F1 scores to increase the weight on high karma comments, 
which gives slightly better performance for the high karma cases but has only a small effect on the macro average. 


\subsection{Baseline and Contrast Systems}

We compare the graph LSTM to a node-independent baseline, which is a feedforward neural network model consisting of input, hidden and softmax layers. This model is a simplification of the graph-LSTM model where there is no connection between nodes. The node-independent model characterizes a comment without reference to either the text of the comment that it is responding to or the comments reacting to it. However, the model does have information on the size of the response subtree via the submission context input features.  Both node-independent and graph-structured models are trained with the same cost function and tuned over the same set of hidden layer dimensions. 

We contrast performance of both architectures with and without using the text of the comment itself. As shown in \cite{fang2016learning}, simply using submission context features (graph, timing, author) gives a strong baseline.
In order to evaluate the role of each direction (forward or backward) in the graph-structured model, we also present results using only the forward direction graph-LSTM for comparison to the bidirectional model. In addition, in order to evaluate the importance of the language of the comment itself vs.\ the language used in the rest of the tree, we perform an interpolation between the graph-LSTM with no language features and the node-independent model with language features.  The relative weight for the two models is tuned on the development set.

\subsection{Karma Level Prediction}

The results for the average F1 scores on the test set are presented in Table \ref{tab:models1}.  In  experiments for all the subreddits, graph-structured models outperform the corresponding node-independent models both with and without language features. Language features also give a greater performance gain when used in the graph-LSTM models. The fact that the forward graph improves over the interpolated models shows that it is not simply the information in the current node that matters for karma of that node.
Finally, while the full model outperforms the forward-only version for all the subreddits, the gain is smaller than that obtained by the forward direction alone over the node-independent model, so the forward direction seems to be more important.



\begin{table}
\begin{center}
\begin{tabular}{|K{1.1cm}|K{0.7cm}|c|c|c|}
\hline \bf Model  &  \bf Text & \bf askwomen & \bf askmen & \bf politics \\ \hline
indep & no  & 53.2 & 48.3 & 46.6 \\ \hline
graph & no  & 54.6 & 52.1 & 47.9 \\ \hline
indep & yes & 52.8 & 50.7 & 47.4 \\ \hline
interp & mix & 54.7 & 52.1 & 48.2 \\ \hline 
graph(f) & yes & 55.0 & 53.3 & 49.9 \\ \hline 
graph & yes & \bf 56.4 & \bf 54.8 & \bf 50.4 \\ \hline 
\end{tabular}
\caption{\label{tab:models1} Average F1 score of karma level prediction for node-independent (indep) vs.\ graph-structured (graph) models with and without text features; interp corresponds to an interpolation of the graph-structured model without text and the node-independent model with text; and graph(f) corresponds to a graph-structured model contains forward direction only. }
\end{center}
\end{table}

\begin{figure*}[t]
\centering
\includegraphics[scale=0.38]{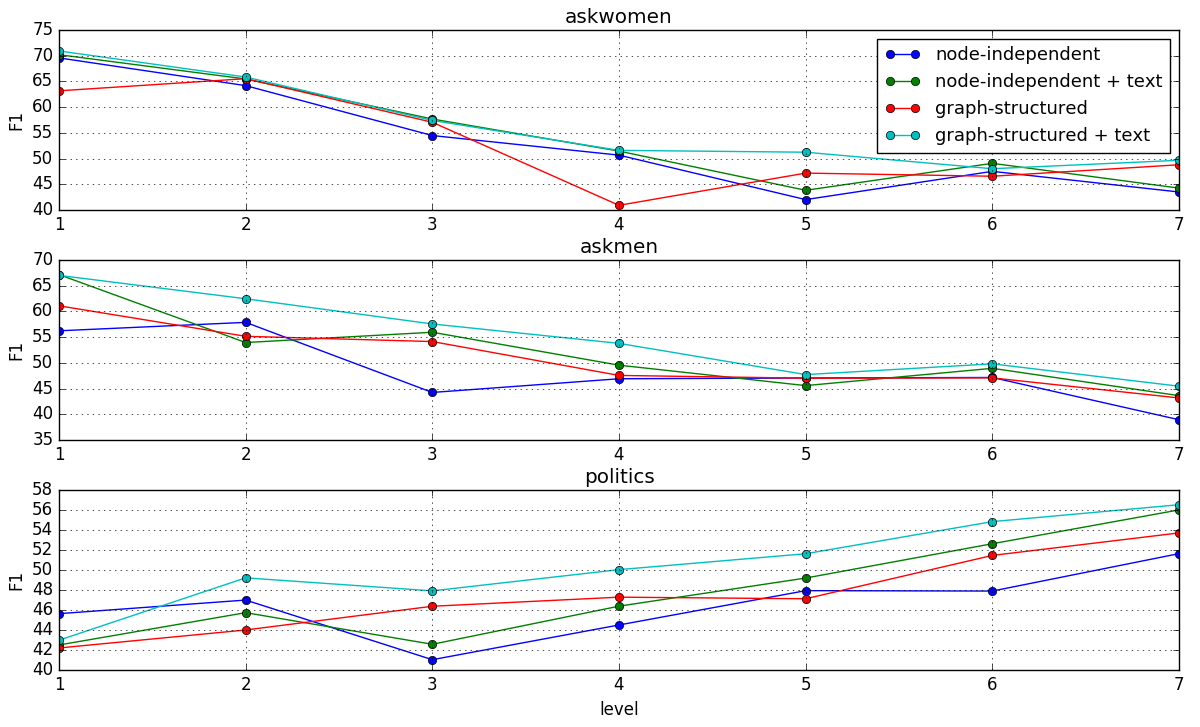}
\caption{F1 scores as a function of the quantized levels for different model configuration.}
\label{fig:levels}
\end{figure*}

The karma prediction results (F1 score) at the different levels is shown in Figure \ref{fig:levels}. While in askmen and askwomen subreddits the overall performance decreases for higher levels, the politics subreddit has an opposite trend. This may be due in part to the lower pruning recall in the politics subreddit, but \cite{fang2016learning} also observe higher performance for high karma levels in the politics subreddit.

\section{Analysis}
\label{sec:analysis}

Here, we present analyses aimed at better understanding the behavior of the graph-structured model and the role of language in prediction. All analyses are performed on the development set. The analyses are motivated by considering possible scenarios that are exceptions to the easy cases, which are: i)~comments that are contributed early in the discussion and spawn large subtrees, likely to have high karma, and ii)~comments with small subtrees that typically have low karma.  
We hypothesized three scenarios where the bidirectional graph-LSTM with text might be useful. One case is controversial comments, which have large subtrees but do not have high karma because of downvotes; these tend to have {\it overprediction} of karma when using only submission context. The other two scenarios involve {\it underprediction} of karma when using only submission context. Early comments associated with few children and a more narrow subtree (see the downward chain in Figure~\ref{fig: visual}) may spawn popular new threads and benefit from the popularity of other comments in the thread (more readers attracted), thus having higher popularity than the number of children suggests. Lastly, comments that are clever or humorous discussion endpoints might have high popularity but small subtrees. These two cases tend to differ in their relative timing in the discussion.




\subsection{Karma Prediction vs.\ Time}





\begin{figure}[t]
\centering
\includegraphics[scale=0.315]{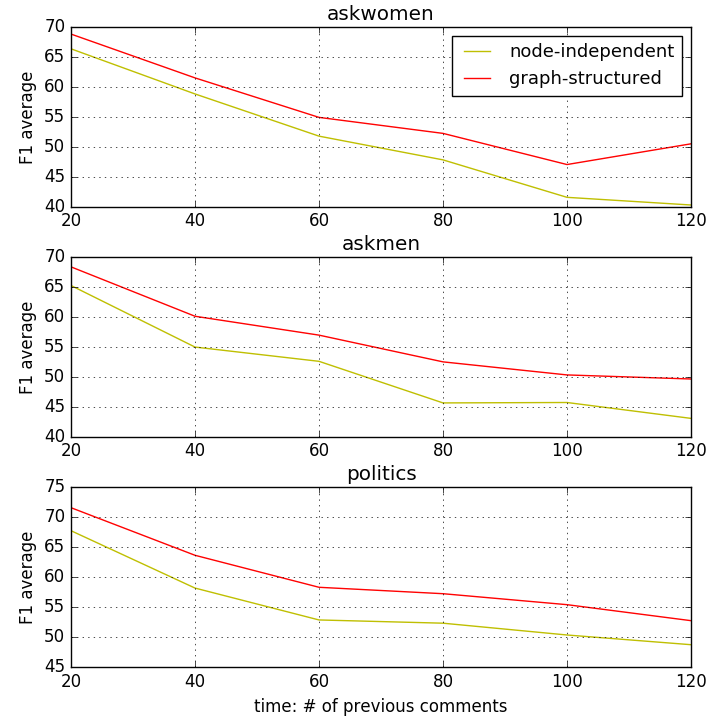}
\caption{Average F1 scores as a function of time, approximated using the number of previous comments quantized in increments of 20.\label{fig:time}}
\end{figure}

The first study looked at where the graph-LSTM provides benefits in terms of timing. We plot the average F1 score as a function of the contribution time in Figure \ref{fig:time}. As an approximation for time, we use the quantized number of comments made prior to the current comment. The plots show that the graph-structured model improves over the node-independent model throughout the discussion. Relative gains are larger towards the end of discussions where the node-independent performance is lower. A similar trend is observed when plotting average F1 as a function of depth in the discussion tree. 

While the use of text in the graph-LSTM seems to help throughout the discussion, we hypothesized that there would be different cases where it might help, and these would occur at different times. Indeed, 93\% of the comments that are overpredicted by more than 2 levels by the node-independent model without text (controversial comments) occur in the first 20\% of the discussion. Comments that are underpredicted by more than 2 occur throughout the discussion and are roughly uniform (13-19\%) over the first half of the discussion, but then quickly ramp down. High-karma comments are rare at the end of the discussion; less than 5\% of the underpredicted comments are in the last 30\%.

\subsection{Importance of Responses}



\begin{figure*}[t]
\includegraphics[scale=0.5]{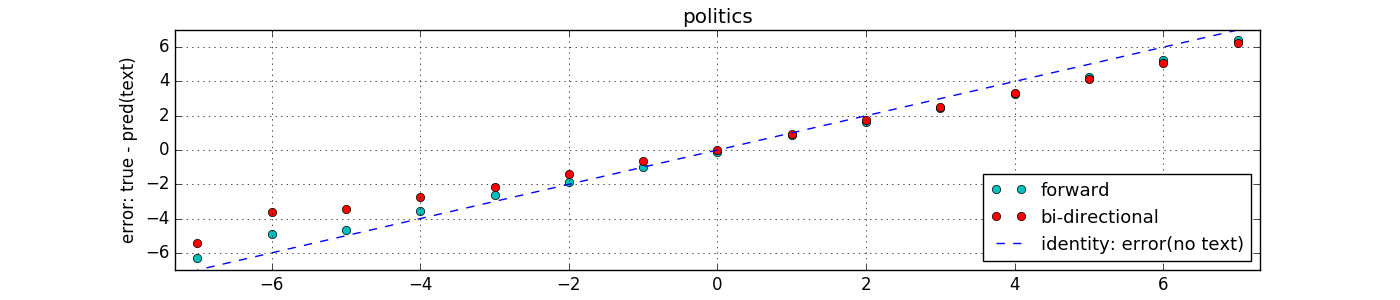}
\caption{The error between the actual karma level and the karma level predicted by the model using both submission context and language features. Negative errors correspond to over-prediction; positive errors correspond to under-prediction.}
\label{fig:lang2}
\end{figure*}

In order to see how the model benefits from using the language cues in \textit{underpredicted} and \textit{overpredicted} scenarios, we look at the size of errors made by the graph-LSTM model with and without text features. In Figure \ref{fig:lang2}, the x-axis indicates the error between the actual karma level and the karma level predicted by the graph-LSTM using submission context features only. The negative errors represent the \textit{overpredicted} comments, and the positive errors represent the \textit{underpredicted} comments. 
The y-axis represents the average error between the actual karma level and the karma level predicted by the model using both submission context and language features.The x=y identity line corresponds to no benefit from language features. Results are presented for the politics subreddit; other subreddits have similar trends but smaller differences for the underpredicted cases. 
 
We compare two models -- bidirectional and forward direction graph-structured LSTM  -- in order to understand the role of the language of the replies vs.\ the comment and its history.
We find that, for the bidirectional graph-LSTM model, language is helping identify overpredicted cases more than underpredicted ones.
The forward direction model also outperforms the node-independent model, but has less benefit in overpredicted cases, consistent with our intuition that controversy is identifiable based on the responses.  Although the comment text input is simply a bag of words, it can capture the mixed sentiment of the responses.

While it is not represented in the plot, larger errors are much less frequent.
Looking at average F1 as a function of the number of children (direct responses), we found that the graph-LSTM mainly benefits nodes that have a small number of children, consistent with the two underprediction scenarios hypothesized.
However, many underpredicted cases are not impacted, since errors due to pruning contribute to 15-40\% of the underpredicted cases, depending on the subreddit (highest for politics). This explains the smaller gains for the positive side of Figure \ref{fig:lang2}.

\subsection{Language Use Analysis}

To provide insights into what the model is learning about language, we looked at individual words associated with different categories of comments, as well as examples of the different error cases.

For the word level analysis, we classified words in two different ways, again using the politics subreddit. First, we associate words in comments with zero or positive karma. For each word in the vocabulary, we calculate the probability of a single-word comment being level zero using
the trained model with a simplified graph structure (a post and a comment) where all the inputs were set to zero except the comment text. 
The lists of positive-karma and zero-karma correspond to the 300  words associated with the lowest and highest probability of zero-karma, respectively. We identified 300 positive-karma and zero-karma reply words in a similar fashion, using a simplified graph with individual words us as inputs for the reply while predicting the comment karma. 

Second, we identified words that may be indicative of comments that are over- and underpredicted by the graph-structured model without text
and for which the graph-LSTM model with text reduced the error by more than 2 levels. Specifically, we choose those words $w$ in comments having the highest ratio $r=p(w|t)/p(w)$, where $t$ indicates an over- or underpredicted comment, subject to minimum occurrence constraints (5 for overpredicted comments, 15 for underpredicted comments). The 50 words with the highest ratio were chosen for each case and any words in both over- and underpredicted sets were eliminated, leaving 47 words. Again, this was repeated for words in replies to over vs.\ underpredicted comments, but with a minimum count threshold of 20, resulting in 45 words.


The lists are noisy, similar to what is often found with topic model, and colored by the language of the subreddit community, but a few trends can be observed.   Looking at the list of words associated with replies to positive-karma comments we noticed  words that indicate humor (``LOL'', ``hilarious''), positive feedback (``Like'', ``Right''), and emotion indicators (``!!'', swearing). 
Words in comments and replies associated with overpredicted (controversial) cases are related to controversial topics (sexual, regulate, liberals), named political parties, and mentions of downvoting or indication that the comment has been edited with the word ``Edit.''

Since the two sets of lists were generated separately, there are words in the over/under-predicted lists that overlap with the zero/non-zero karma lists (12 in the reply lists, 20 in the comment lists). 
The majority of the overlap (26/32 words) is consistent with the intuition that words on the underpredicted list should be associated with positive-karma, and words on the overpredicted list might overlap with the zero-karma list.

\begin{figure}[t]
\centering
  \includegraphics[scale=0.32]{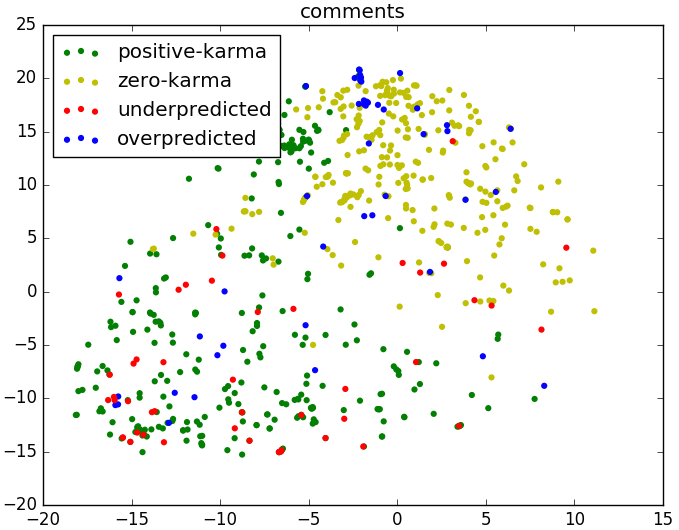}%
\caption{The mapping of the words in the comments to the shared space using t-SNE in politics subreddit. Shown are the words that are highly associated with positive-karma, negative-karma, underpredicted and overpredicted comments.}
\label{fig:emb}
\end{figure}

Rather than providing word lists, many neural network studies illustrate trends using word embedding visualization. 
The embeddings of the words from the union of lists for positive-karma, zero-karma, underpredicted and overpredicted comments and replies were together used to learn a t-SNE mapping. The results are plotted for comments in Figure ~\ref{fig:emb}, which shows that the words that are associated with underpredicted comments (red) are aligned with positive-karma words (green) for both comment text and text in replies. Words associated with overpredicted comments (blue) are more scattered, but they  are somewhat more like the zero-karma words (yellow). The trends for words in replies are similar.

Table~\ref{tab:examples} lists examples of the different error scenarios with the reference karma and predictions of different models (node-independent without text, feedforward graph-LSTM with text, and the full biLSTM). The first two examples are overpredicted (controversial) cases, where ignoring text leads to a high karma prediction, but the reference is zero. In the first case, the forward model incorrectly predicts high karma because ``Republican'' tends to be associated with positive karma. The model leveraging reply text correctly predicts the low karma. In the second case, the forward model captures reduces the prediction, but again having the replies is more helpful. The next two cases are examples of underprediction due to small subtrees. Example 3 is incorrectly labeled as level 0 by the forward and no-text models, but because the responses mention ``nice joke'' and ``accurate analogy,'' the bidirectional model is able to identify it as level 7. Example 4 has only one child, but both  models using language correctly predict level 7, probably because the model has learned that references to ``Colbert'' are popular. The next two examples are underpredicted cases from early in the discussion, many of which expressed an opinion that in some way provided multiple perspectives. Finally, the last two examples represent instances where neither model successfully identifies a high karma comment, which often involve analogies. Unlike the ``titanic'' analogy, these did not have sufficient cues in the replies.

\begin{table*}
\begin{center}
{\small
\begin{tabular}{|c|cccc|p{5in}|}\hline
Ex & \multicolumn{4}{c|}{karma} & Comment\\ \hline
1 & 0 & 7 & 7 & 0 & Republicans are fundamentally dishonest. (politics, id:1x9pcx) \\ \hline
2 & 0 & 7 & 4 & 0 & That is rape. She was drunk and could not consent. Period. Any of the supposed ”evidence” otherwise is nothing but victim blaming. (askwomen, id:2h8pyh) \\ \hline
3 & 7 & 0 & 0 & 7 & The liberals keep saying the titanic is sinking  but my side is 500 feet in the air. (politics, id:1upfgl) \\ \hline
4 & 7 & 3 & 7 & 7 & I miss your show, Stephen Colbert. (askmen, id:2qmpzm)\\ \hline
5 & 7 & 3 & 7 & 7 &  that is terrifying.  they were given the orders to bust down the door without notice  to the residents, thereby placing themselves in danger.  and ultimately, placing the lives of the residents in danger (who would be acting out of fear and self-defense) (politics, id:1wzwg6)\\ \hline
6 & 7 & 0 & 5 & 6 & It's something, and also would change the way that Police unions and State Prosecutors work. I don't fundamentally agree with the move, since it still necessitates abuse by the State, but it's something. (politics, id:27chxr)\\ \hline
7 & 6 & 0 & 0 & 0 &  Chickenhawks always talk a big game as long as someone else is doing the fighting. (politics, id:1wbgpd)\\ \hline
8 & 6 & 0 & 0 & 0 & $[$They$]$ use statistics in the same way that a drunk uses lampposts: for support, rather than illumination.  -Andrew Lang. (politics, id:1yc2fj)\\ \hline
\end{tabular}
}
\end{center}
\caption{Example comments and karma level predictions: reference, no text, graph(f), graph. \label{tab:examples}}
\end{table*}

\section{Related Work}
\label{sec:literature}

The problem of predicting popularity in social media platforms has been the subject of several studies. Popularity as defined in terms of volume of response has been explored for shares on Facebook \cite{cheng2014can} and Twitter \cite{bandari2012pulse} and Twitter retweets \cite{tan2014effect,zhao2015seismic,bi2016modeling}. Studies on Reddit predict karma
as popularity
 \cite{lakkaraju2013s,jaech2015talking,He2016} or as community endorsement \cite{fang2016learning}.
Popularity prediction is a difficult task where many factors can play a role, which is why most prior studies control for specific factors, including topic \cite{tan2014effect,weninger2013exploration}, timing \cite{tan2014effect,jaech2015talking}, and/or comment content \cite{lakkaraju2013s}.
Controlling for specific factors is useful in understanding the components of a successful post, but it does not reflect a realistic scenario. Studies that do not include such constraints have looked at Twitter retweets \cite{bi2016modeling} and Reddit karma \cite{He2016,fang2016learning}.  

The work in \cite{He2016} uses reinforcement learning to identify popular threads to track given the past comment history, so it is learning language cues relevant to high karma but it does not explicitly predict karma. In addition, it models relevance via an inner-product of past and new comment embeddings, and uses an LSTM to model inter-comment dependencies among a collection of comments irrespective of their sibling-parent relationship, whereas the LSTM in our work is over a graph that accounts for this relationship.

The work most closely related to our study is \cite{fang2016learning}. The node-independent baseline implemented in our study is equivalent to their feedforward network baseline, but the results are not directly comparable because of differences in training (we use more data) and input features. The most important difference in our approach is the representation of textual context using a bidirectional graph-LSTM, including the history behind and responses to a comment. Other differences are: i) Fang {\it et al.}\ use an LSTM to characterize comments, while our model uses a simple bag-of-words approach, and ii) they learn latent submission context models to determine the relative importance of textual cues, while our approach uses a submission context SVM to prune low karma comments (ignoring their text). Allowing for differences in baselines, we note that the absolute gain in performance from using text features is larger for our model, which represents language context. 

Tree LSTMs are a modification of sequential LSTMs that have been proposed for a variety of sentence-level NLP tasks \cite{tai2015improved,zhu2015long,zhang2016top,lezuidema2015}. 
The architecture of tree LSTMs varies depending on the task. Some options include summarizing over the children, adding a separate forget gate for each child \cite{tai2015improved}, recurrent propagation among siblings \cite{zhang2016top}, or use of stack LSTMs \cite{Dyer+16}. Our work differs from these studies in two respects: the tree structure here characterizes a discussion rather than a single sentence; and our architecture incorporates both hierarchical and temporal recursions in one LSTM unit.

\section{Conclusion}

In summary, this paper presents a novel approach for modeling threaded discussions on social media using a graph-structured bidirectional LSTM which represents both hierarchical and temporal conversation structure. The propagation of hidden state information in the graph provides a mechanism for representing contextual language, including the history that a comment is responding to as well as the ensuing discussion it spawns. Experiments on Reddit discussions show that the graph-structured LSTM leads to improved results in predicting comment popularity compared to a node-independent model. Analyses show that the model benefits prediction over the extent of the discussion, and that language cues are particularly important for distinguishing controversial comments from those that are very positively received. Responses from even a small number of comments seem to be useful, so it is likely that the bidirectional model would still be useful with a short-time lookahead for early prediction of popularity.

While we evaluate the model on predicting the popularity of comments in specific forums on Reddit, it can be applied to other social media platforms that maintain a threaded structure or possibly to citation networks. In addition to popularity prediction, we expect the model would be useful for other tasks for which the responses to comments are informative, such as detecting topic or opinion shift, influence or trolls. With the more fine-grained feedback increasingly available on social media platforms (e.g.\ laughter, love, anger, tears), it may be possible to distinguish different types of popularity as well as levels, e.g. shared sentiment vs.\ humor.

In this study, the model uses a simple bag-of-words representation of the text in a comment; more sophisticated attention-based models and/or feature engineering may improve performance. In addition, performance of the model on underpredicted comments appears to be limited by the pruning mechanism that we introduced. It would be useful to explore the tradeoffs of reducing the amount of pruning vs.\ using a more complex classifier for pruning. Finally, it would be useful to evaluate performance using a short window lookahead for responses, rather than the full discussion tree.

\section*{Acknowledgments}

Omitted for blind review.

\bibliography{graph_lstm}
\bibliographystyle{acl2012}

\end{document}